\documentclass[10pt,twocolumn,letterpaper]{article}

\usepackage[pagenumbers]{cvpr} 

\definecolor{cvprblue}{rgb}{0.21,0.49,0.74}
\usepackage[pagebackref,breaklinks,colorlinks,allcolors=cvprblue]{hyperref}

\def\paperID{*****} 
\def\confName{CVPR}
\def\confYear{2025}

\title{United we stand, Divided we fall: Handling Weak Complementary Relationships for Audio-Visual Emotion Recognition in Valence-Arousal Space}

\author{R. Gnana Praveen, Jahangir Alam, Eric Charton\\
Computer Research Institute of Montreal (CRIM), Canada\\
{\tt\small gnana-praveen.rajasekhar@crim.ca, jahangir.alam@crim.ca,eric.charton@crim.ca}
}

\begin{document}
\maketitle
\begin{abstract}
Audio and visual modalities are two predominant contact-free channels in videos, which are often expected to carry a complementary relationship with each other. However, they may not always complement each other, resulting in poor audio-visual feature representations. 
In this paper, we introduce Gated Recursive Joint Cross Attention (GRJCA) using a gating mechanism that can adaptively choose the most relevant features to effectively capture the synergic relationships across audio and visual modalities.  
Specifically, we improve the performance of Recursive Joint Cross-Attention (RJCA) by introducing a gating mechanism to control the flow of information between the input features and the attended features of multiple iterations depending on the strength of their complementary relationship. For instance, if the modalities exhibit strong complementary relationships, the gating mechanism emphasizes cross-attended features, otherwise non-attended features. To further improve the performance of the system, we also explored a hierarchical gating approach by introducing a gating mechanism at every iteration, followed by high-level gating across the gated outputs of each iteration. The proposed approach improves the performance of RJCA model by adding more flexibility to deal with weak complementary relationships across audio and visual modalities. Extensive experiments are conducted on the challenging Affwild2 dataset to demonstrate the robustness of the proposed approach. 
By effectively handling the weak complementary relationships across the audio and visual modalities,  
the proposed model achieves a Concordance Correlation Coefficient (CCC) of 0.561 (0.623) and 0.620 (0.660) for valence and arousal respectively on the test set (validation set). This shows a remarkable improvement over the baseline of 0.211 (0.240) and 0.191 (0.200) for valence and arousal, respectively, in the test set (validation set), achieving competitive performance 
in the valence-arousal challenge of the 8th Affective Behavior Analysis in-the-Wild (ABAW) competition. 


\end{abstract}    
\section{Introduction}
\label{sec:intro}

With the advancement of technology, empowering machines with affective capabilities is a fundamental requirement for Human-Computer Interaction (HCI). It has a lot of applications in a wide range of environments such as health-care \cite{RAJASEKHAR2021104167}, autonomous driving \cite{CHEN2022104569}, e-learning \cite{10.1145/3385209.3385231}, etc. Automatic recognition of human emotions is a challenging problem as the expressions pertinent to the emotional states are quite complex and extremely diverse across individuals and cultures \cite{Anagnostopoulos}. Emotion Recognition (ER) can be formulated as the problem of classification or regression of emotions. 
In recent years, regression of expressions is gaining attention as it has the potential to capture a wide range of expressions, which can be useful in many applications such as estimation of pain intensity levels in the healthcare domain \cite{9320216}, engagement intensity levels in business or educational settings \cite{8615851}, etc. 
Depending on the granularity of labels, the regression of emotions can be formulated as ordinal regression or continuous regression. Compared to ordinal regression, continuous (or dimensional) regression is even more challenging due to the complex process of obtaining annotations in continuous dimensions. Valence and arousal are widely used dimensions to estimate emotion intensities in a continuous domain. 
Valence reflects the wide range of emotions in the dimension of pleasantness, from being negative (sad) to positive (happy). In contrast, arousal spans a range of intensities from passive (sleepiness) to active (high excitement). 

Human emotions can be often conveyed through various modalities such as face, voice, text, physiology, etc. Out of all the modalities, facial and vocal expressions are the widely explored modalities in videos for multimodal ER \cite{10.1145/2682899}. 
Cross Attention (CA) has been widely explored to capture the complementary relationships across audio and visual modalities in several applications such as event localization \cite{9423042}, action localization \cite{lee2021crossattentional}, and emotion recognition \cite{9667055}. 
Recently, sophisticated cross attention models have been proposed by introducing
joint feature representation in the cross attentional framework to enhance the feature representations by simultaneously capturing intra- and inter-modal relationships \cite{Praveen_2022_CVPR,10005783,praveen2023recursive}. Even though these models show promising performance, they rely on the assumption that audio and visual modalities always exhibit complementary relationship.  
However, the audio and visual modalities may not always be complementary to each other, they may also conflict with each other \cite{9552921}. 
Praveen et al. \cite{10584250} investigated the problem of weak complementary relationships for ER with visual analysis and showed that weak complementary relationships degrade the cross-attended features, resulting in poor audio-visual feature representations. 
To tackle this problem, Praveen et al. \cite{10687371} proposed a Dynamic Cross Attention (DCA) model to adaptively choose the most relevant features among the cross attended and unattended features based on gating mechanism. In this work, we further extend their approach on the recently proposed RJCA \cite{praveen2023recursive} model by introducing the gating mechanism to control the flow of information across the original features and cross attended features of every iteration. By controlling the flow of information across the features of all iterations of RJCA model, we have more flexibility to capture the most relevant information across all the iterations. To further improve the system, we also explored hierarchical gating, where the gating mechanism is employed in every iteration, followed by high-level gating mechanism across the gated outputs of every iteration. The proposed approach demonstrates superior performance over the DCA \cite{10584250} model as it can leverage the attended features of all iterations instead of only the final attended features.   
The major contributions of the paper are as follows. 
\begin{itemize}
    \item We improve the performance of RJCA model by handling weak complementary relationships across the audio and visual modalities by controlling the flow of information across the features of all iterations and original features.
    \item The proposed model has been further enhanced by introducing a hierarchical gating mechanism to capture more refined semantic features. 
    \item A detailed set of experiments are conducted to evaluate the contribution of the proposed model on the Affwild2 dataset, achieving better generalization ability and superior performance over prior state-of-the-art models.
\end{itemize}
\section{Related Work}

\subsection{A-V fusion for dimensional ER}


Atmaja et al \cite{9052916} proposed a weighted summation of the loss functions in a multi-task learning framework, where the weights of the individual loss components are adjusted to improve the performance. 
Enrique et al also \cite{10.1145/2512530.2512534} explored the multi-stage fusion of predictions obtained from the multiple features of each modality in a hierarchical fashion. Haifeng et al \cite{10.1145/3347320.3357690} extracted deep spatio-temporal feature vectors of the images and spectrograms by combining pre-trained 2D-CNNs with 1D-CNNs and also exploited spatial-temporal graph convolutional network (STGCN) to obtain the geometric features based on facial landmarks. They have employed hybrid fusion using BLSTMs to obtain final predictions. Mihalis et al \cite{10.1145/2502081.2502201} studied the correlation across the emotional dimensions and found that each emotional dimension is more correlated with other emotional dimensions than with the features of individual modalities. Ehab et al \cite{10.1145/3242969.3242972} investigated the optimal trade-off between continuous and discrete emotion representations and found that jointly modeling the discrete and continuous emotion representations yields better performance. Though the above-mentioned approaches have shown improvement in dimensional ER, they fail to capture the inter-modal relationships and relevant salient features specific to the task. 


\subsection{Attention for A-V fusion}
Huang et al \cite{9053762} explored multimodal transformers to obtain effective multimodal feature representations by latently adapting audio to visual modality into a common feature space via cross-modal transformers. Srinivas et al \citep{srini_2021_SLT} also explored multimodal transformers, where the cross-attention module is integrated with the self-attention module to obtain the A-V cross-modal feature representations. Tran et al \cite{9747278} investigated the potential of using pre-trained audio-visual transformers for ER and showed that fine-tuning pre-trained models improves performance. Praveen et al \cite{10005783} proposed a cross-attention model to effectively capture the complementary relationships across the audio and visual modalities. They further extended their approach by introducing joint feature representations in the cross-attentional framework \cite{Praveen_2022_CVPR,10005783} and recursive attention \cite{praveen2023recursive,Praveen_2024_CVPR}. Although these sophisticated cross attention models have shown promising performance, they often rely on the assumption that audio and visual modalities always exhibit strong complementary relationships. When the audio and visual modalities exhibit weak complementary relationship, they fail to retain their superior performance. 

\subsection{Gating-Based Attention for A-V Fusion}

Gating mechanisms have been explored for multimodal fusion to control the importance of each modality in the fusion mechanism \cite{10.1007/s00521-019-04559-1}. Decky et al \cite{9726856} proposed a gated-sequence neural network for dimensional ER, where the gating mechanism is explored along with simple feature concatenation to adaptively fuse the modalities based on their relative importance. Lee et al \cite{9706879} addressed the problem of noisy or corrupted modalities using a leaky gated cross attention by adaptively emphasizing the salient features of A and V modalities for action localization. Ayush et al \cite{9053012} explored gated cross-attention along with self-attention 
for sentiment analysis and showed that exploiting the gating mechanism using a nonlinear transformation with cross-attention helps to control the impact of imperfect modalities. 
To deal with complementary relationships, Praveen et al. \cite{10687371,10584250} proposed Dynamic Cross Attention (DCA) by employing a gating mechanism to choose the relevant cross-attended or unattended features based on the strength of the complementary relationship. However, they rely only on the final attended features and original features, which may not capture the relevant information available in the features of individual iterations, resulting in suboptimal performance. In this work, we further extend the idea of DCA by employing the gating mechanism over the features of all the iterations of RJCA model. 

\section{Proposed Approach}
In this section, we briefly introduce our baseline fusion model, RJCA, as a preliminary followed by the proposed approach. 
\begin{figure*}[!t]
\centering
\includegraphics[width=1.0\linewidth]{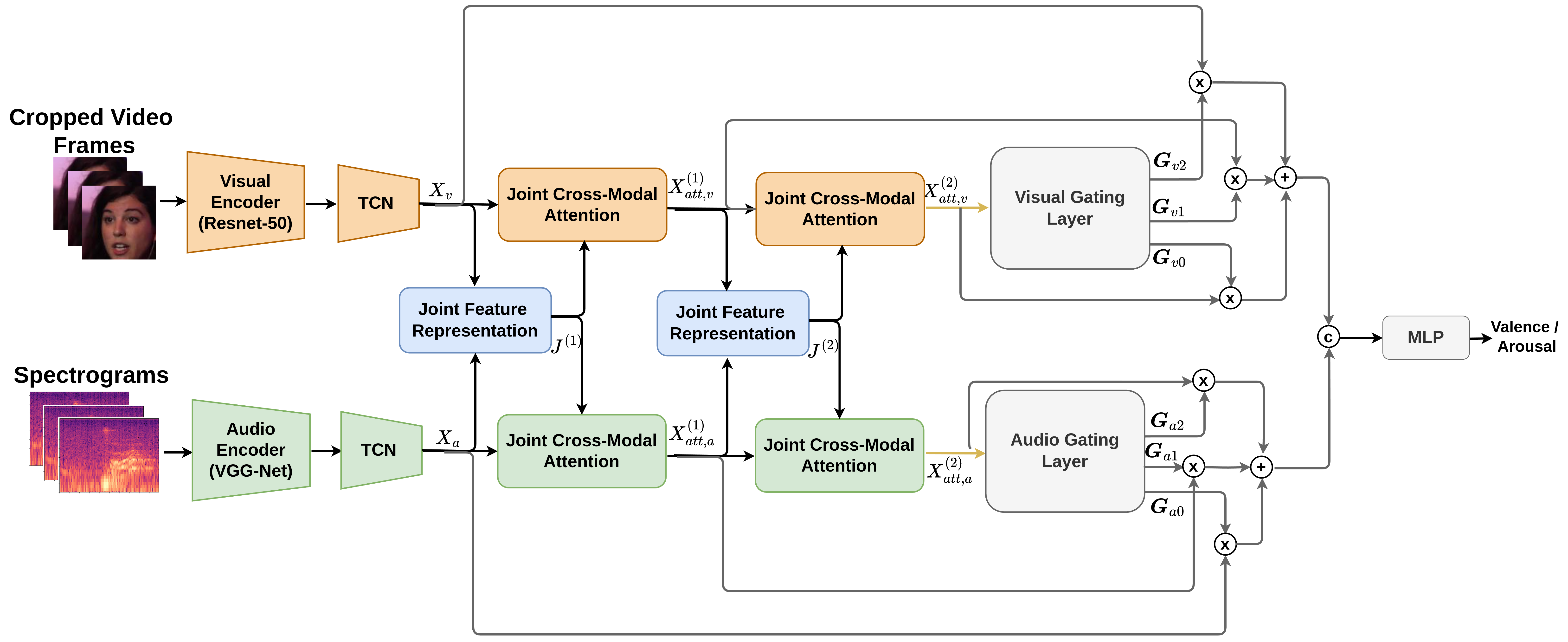}
\caption{Illustration of the proposed GRJCA model for two iterations. Different colorized arrows are used to denote the gating mechanism. Best viewed in color.}
\label{VA}
\end{figure*}

\subsection{Notations} 
Given the video subsequence $S$ of $L$ frames, the feature representations of audio and visual modalities are denoted as 
${\boldsymbol X}_{a} =  \{ \boldsymbol x_{a}^1, \boldsymbol x_{a}^2, ..., \boldsymbol x_{a}^L \boldsymbol \} \in \mathbb{R}^{d_a\times L}$,
${\boldsymbol X}_{v} =  \{ \boldsymbol x_{v}^1, \boldsymbol x_{v}^2, ..., \boldsymbol x_{v}^L \boldsymbol \} \in \mathbb{R}^{d_v\times L}$, respectively, where $d_a$ and $d_v$ are dimensions of audio and visual modalities. $\boldsymbol x_{a}^l$ and $\boldsymbol x_{v}^l$ represent the feature vectors of the individual corresponding frames of audio and visual modalities. The joint feature representation ($\boldsymbol J$) is obtained by concatenating the feature vectors of audio and visual modalities, followed by fully connected layer, which is given by
$ {\boldsymbol J = FC([{\boldsymbol X}_{a} ; {\boldsymbol X}_{v}]) \in\mathbb{R}^{d\times L}}$
where $d = {d_a} + {d_v}$ denotes the dimensionality of $\boldsymbol J$ and $FC$ denote fully connected layer. 

\subsection{Recursive Joint Cross Attention (RJCA)}
To capture the semantic relevance across and within the modalities, cross-correlation is computed between the joint feature representation $\boldsymbol J$ and the feature representations of individual modalities (${\boldsymbol X}_{a}$ and ${\boldsymbol X}_{v}$), which is given by   
\begin{align}
  \boldsymbol C_{a}= \tanh \left(\frac{{\boldsymbol X}_{ a}^\top{\boldsymbol W}_{ja}{\boldsymbol J}}{\sqrt d}\right)
\\
  \boldsymbol C_{v}= \tanh \left(\frac{{\boldsymbol X}_{ v}^\top{\boldsymbol W}_{jv}{\boldsymbol J}}{\sqrt d}\right)
\end{align}
where $\boldsymbol C_{a}$ and $\boldsymbol C_{v}$ represent the joint cross-correlation matrices, ${\boldsymbol W}_{ja} \in\mathbb{R}^{d_a\times d}$, ${\boldsymbol W}_{jv} \in\mathbb{R}^{d_v\times d}$ represents learnable weight matrices of audio and visual modalities respectively. 

Now the joint cross-correlation matrices are used to compute the attention maps of the individual modalities as 
\begin{align}
\boldsymbol H_{a}=ReLU(\boldsymbol X_{a} \boldsymbol W_{ca} {\boldsymbol C}_a)
\\
\boldsymbol H_{v}=ReLU(\boldsymbol X_{v} \boldsymbol W_{cv} {\boldsymbol C}_v)
\end{align}

where ${\boldsymbol W}_{ca} \in\mathbb{R}^{{L}\times {L}}$, ${\boldsymbol W}_{cv} \in\mathbb{R}^{{L}\times {L}}$ denote learnable weight matrices for audio and visual modalities. 

The attention maps are further used to compute the attended features of the individual modalities as: 
\begin{align}
{\boldsymbol X}_{att, a} = \boldsymbol H_{a} \boldsymbol W_{ha} + \boldsymbol X_{a}
\\
{\boldsymbol X}_{att, v} = \boldsymbol H_{v} \boldsymbol W_{hv} + \boldsymbol X_{v}  
\end{align}
where $\boldsymbol W_{ha} \in\mathbb{R}^{{L}\times {L}}$ and $\boldsymbol W_{hv} \in\mathbb{R}^{{L}\times {L}}$  denote the learnable weight matrices for audio and visual respectively. 

To obtain more refined feature representations, the attended features of the individual modalities are again fed as input to the joint cross-attention fusion model in a recursive fashion, which is given by  
\begin{align}
{\boldsymbol X}_{att, a}^{(t)} = \boldsymbol H_{a}^{(t)} \boldsymbol W_{ha}^{(t)} + \boldsymbol X_{a}^{(t-1)}
\\
{\boldsymbol X}_{att,v}^{(t)} = \boldsymbol H_{v}^{(t)} \boldsymbol W_{hv}^{(t)} + \boldsymbol X_{v}^{(t-1)}  
\end{align}

where $\boldsymbol W_{ha}^{(t)} \in\mathbb{R}^{{L}\times {L}}$, $\boldsymbol W_{hv}^{(t)} \in\mathbb{R}^{{L}\times {L}}$ denote the learnable weight matrices of audio and visual modalities, respectively, and $t$ refers to the recursive step. 

The attended features of the individual modalities at the last iteration $t_e$ are concatenated to obtain the multimodal feature representation ${\boldsymbol X}_{att}$. 

\subsection{Gated Recursive Joint Cross Attention (GRJCA)}
Though audio and visual modalities are expected to be complementary to each other \cite{9758834}, they may not always complement each other \cite{9552921}. 
Therefore, we present a gated attention mechanism to adaptively fuse the audio and visual features, based on the compatibility of the modalities with each other i.e., the gated attention emphasizes the cross-attended features when the features exhibit a strong complementary relationship, otherwise non-attended features. Specifically, we employ the gating mechanism to control the flow of information across the original features and the attended features of all iterations to leverage the semantic information across all the iterations. By adaptively emphasizing the semantic features across all the iterations and the original unattended features, the proposed model is able to effectively handle the weak complementary relationships across audio and visual modalities.  
Given the attended features of multiple iterations and non-attended features from audio and visual modalities, we design a gating layer using a fully connected layer for every iteration of each modality separately to obtain the attention weights for the attended and non-attended features, which are given by 
\begin{align}
{\boldsymbol W}_{go,v} = {\boldsymbol X}_{att,v}^{(t_e)} \boldsymbol W_{gl,v}
\\
{\boldsymbol W}_{go,a} = {\boldsymbol X}_{att,a}^{(t_e)} \boldsymbol W_{gl,a}
\end{align}
where $\boldsymbol W_{gl,a} \in\mathbb{R}^{d_a\times (M+1)}$, $\boldsymbol W_{gl,v}\in\mathbb{R}^{d_v\times (M+1)}$ are the learnable weights of the gating layers and $\boldsymbol W_{go,a}^{(t)} \in\mathbb{R}^{L\times (M+1)}$, $\boldsymbol W_{go,v}^{(t)} \in\mathbb{R}^{L\times (M+1)}$ are outputs of gating layers of audio and visual modalities respectively. The number of output units of the gating layer is determined by the number of iterations and the original unattended features i.e., $(M+1)$ where $M$ denotes the total number of iterations. To obtain probabilistic attention scores, the output of the gating layers is activated using a softmax function with a small temperature $T$, as given by 
\begin{align}
{\boldsymbol G}_{a}
=\frac{ e^{{\boldsymbol W}_{go,a}/T}}{\overset{ K}{\underset{ k\boldsymbol=1}{\sum}} e^{{\boldsymbol W}_{go,a}/T}}    
\quad  \text{and} \quad
{\boldsymbol G}_{v}
=\frac{ e^{{\boldsymbol W}_{go,v}/T}}{\overset{K}{\underset{ k\boldsymbol=1}{\sum}} e^{{\boldsymbol W}_{go,v}/T}}    
\end{align}
where $\boldsymbol {G}_{a} \in\mathbb{R}^{L\times (M+1)}, \boldsymbol {G}_{v}\in\mathbb{R}^{L\times (M+1)}$ denotes the probabilistic attention scores of audio and visual modalities. $K$ denotes the number of output units of the gating layer, which is $M+1$. In our experiments, we have empirically set the value of $T$ to 0.1. 

\begin{figure*}[!t]
\centering
\includegraphics[width=1.0\linewidth]{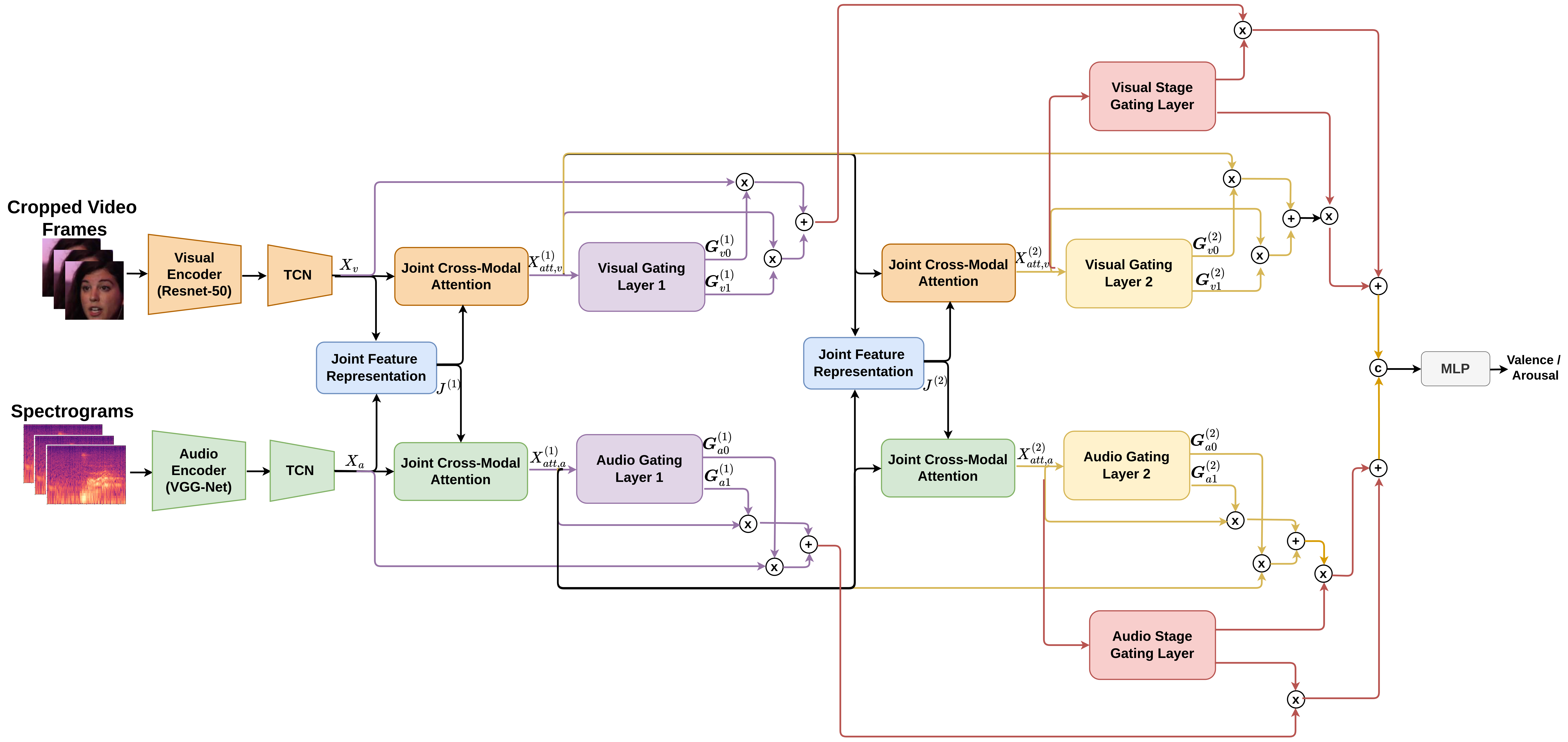}
\caption{Illustration of the proposed HGRJCA model with two iterations. Different colorized arrows are used to denote different gating layers. Best viewed in color.}
\label{VA}
\end{figure*}
The probabilistic attention scores of the gating layer help to estimate the semantic relevance of attended features of every iteration and the original features. 
The columns of $\boldsymbol G_a$ correspond to the probabilistic attention scores of each iteration $t$ and the original unattended features. To multiply with the corresponding feature vectors, each column is replicated to match the dimension of the corresponding feature vectors. 
Now, the replicated attention scores  
are multiplied with the corresponding features of the respective modalities, which is further fed to the ReLU activation function as: 
\begin{align}
{\boldsymbol X}_{att,ga} = \text{ReLU}({\overset{M}{\underset{ t\boldsymbol=0}{\sum}} {\boldsymbol X}_{att,ga}^{(t)}} \otimes \boldsymbol G_{at}) \\
{\boldsymbol X}_{att,gv} = \text{ReLU}({\overset{M}{\underset{ t\boldsymbol=0}{\sum}} {\boldsymbol X}_{att,gv}^{(t)}} \otimes \boldsymbol G_{vt})
\end{align}
where $\otimes$ denotes element-wise multiplication. Here $t=0$ represents the original unattended features. ${\boldsymbol X}_{att,ga}$ and ${\boldsymbol X}_{att,gv}$ denote the final attended features of audio and visual modalities, which is concatenated and fed to Multi Layer Perceptron (MLP) for predicting valence and arousal. 

\subsection{Hierarchical Gated Recursive Joint Cross Attention (HGRJCA)}
We also explored hierarchical gating approach, where the gating mechanism is employed at every iteration, followed by a high level gating mechanism to control the flow of information over the gated outputs of individual iteration. Given the attended features of multiple iterations and non-attended features from audio and visual modalities, we design a gating layer for every iteration of each modality separately, which is given by 
\begin{align}
{\boldsymbol W}_{go,v}^{(t)} = {\boldsymbol X}_{att,v}^{(t)} \boldsymbol W_{gl,v}^{(t)}
\\
{\boldsymbol W}_{go,a}^{(t)} = {\boldsymbol X}_{att,a}^{(t)} \boldsymbol W_{gl,a}^{(t)}
\end{align}
where $\boldsymbol W_{gl,a}^{(t)} \in\mathbb{R}^{d_a\times 2}$, $\boldsymbol W_{gl,v}^{(t)}\in\mathbb{R}^{d_v\times 2}$ are the learnable weights of the gating layers of $t$-th iteration and $\boldsymbol W_{go,a}^{(t)} \in\mathbb{R}^{L\times 2}$, $\boldsymbol W_{go,v}^{(t)} \in\mathbb{R}^{L\times 2}$ are outputs of gating layers of audio and visual modalities of $t$-th iteration respectively. The probabilistic attention scores of the outputs of the gating layers is obtained using a softmax function with a small temperature $T$, as given by 
\begin{align}
{\boldsymbol G}_{a}^{(t)}
=\frac{ e^{{\boldsymbol W}_{go,a}^{(t)}/T}}{\overset{ K}{\underset{ k\boldsymbol=1}{\sum}} e^{{\boldsymbol W}_{go,a}^{(t)}/T}}    
\quad  \text{and} \quad
{\boldsymbol G}_{v}^{(t)}
=\frac{ e^{{\boldsymbol W}_{go,v}^{(t)}/T}}{\overset{K}{\underset{ k\boldsymbol=1}{\sum}} e^{{\boldsymbol W}_{go,v}^{(t)}/T}}    
\end{align}
where $\boldsymbol {G}_{a}^{(t)} \in\mathbb{R}^{L\times 2}, \boldsymbol {G}_{v}^{(t)}\in\mathbb{R}^{L\times 2}$ denotes the probabilistic attention scores of audio and visual modalities. $K$ denotes the number of output units of the gating layer, which is $2$. Note that the two output units correspond to the input and output features of the fusion model at the $t^{th}$ iteration. 

The two columns of $\boldsymbol G_a^{(t)}$ correspond to the probabilistic attention scores of iteration $t$ (first column) and iteration $t-1$ (second column). Similar to GRJCA, each column is replicated to match the dimension of the corresponding feature vectors, which is denoted by $\boldsymbol G_{a0}^{(t)}$, $\boldsymbol G_{a1}^{(t)}$ and $\boldsymbol G_{v0}^{(t)}$, $\boldsymbol G_{v1}^{(t)}$ for audio and visual modalities respectively. Now, the replicated attention scores  
are multiplied with the corresponding features of iteration $t$ and $(t-1)$ of the respective modalities, which is further fed to the ReLU activation function as: 
\begin{align}
{\boldsymbol X}_{att,gv}^{(t)} = \text{ReLU}({\boldsymbol X}_{att,v}^{(t-1)} \otimes \boldsymbol G_{v0}^{(t)} + {\boldsymbol X}_{att,v}^{(t)} \otimes \boldsymbol G_{v1}^{(t)}) \\ 
{\boldsymbol X}_{att,ga}^{(t)} = \text{ReLU}({\boldsymbol X}_{att,a}^{(t-1)} \otimes \boldsymbol G_{a0}^{(t)} + {\boldsymbol X}_{att,a}^{(t)} \otimes \boldsymbol G_{a1}^{(t)})  
\end{align}
where $\otimes$ denotes element-wise multiplication. ${\boldsymbol X}_{att,ga}^{(t)}$ and ${\boldsymbol X}_{att,gv}^{(t)}$ denote the attended features of audio and visual modalities of iteration $t$ respectively. 

To further enhance the flow of information across all the iterations of the RJCA model, we employ a high-level gating mechanism that can obtain semantic features by controlling the flow of information across the gated features of all iterations. Given the features obtained from the gating layers of every iteration, the final features for audio and visual modalities can be obtained as 
\begin{align}
{\boldsymbol X}_{att,ga}^{(f)} = \text{ReLU}({\overset{M}{\underset{ t\boldsymbol=1}{\sum}} {\boldsymbol X}_{att,ga}^{(t)}} \otimes \boldsymbol G_{vt}^{(f)}) \\
{\boldsymbol X}_{att,gv}^{(f)} = \text{ReLU}({\overset{M}{\underset{ t\boldsymbol=1}{\sum}} {\boldsymbol X}_{att,gv}^{(t)}} \otimes \boldsymbol G_{vt}^{(f)})
\end{align}
where $M$ denotes the number of iterations, ${\boldsymbol X}_{att,ga}^{(f)}$ and ${\boldsymbol X}_{att,gv}^{(f)}$ denote the final attended features of audio and visual modalities respectively, $\boldsymbol G_{at}^{(f)}$ and $\boldsymbol G_{vt}^{(f)}$ denote the gating outputs of the final gating layer. 

\section{Experimental Setup}

\subsection{Datasets}
Affwild2 is the largest dataset in the field of affective computing, consisting of $594$ videos collected from YouTube, all captured in-the-wild \cite{Kollias}. 
The dataset has been extensively used to conduct a series of challenges on affective analysis including, expression recognition, action unit recognition, valence and arousal estimation \cite{zafeiriou2017aff,kollias2020analysing,kollias2021analysing,kollias2022abaw,kollias2023abaw,kollias2023abaw2,kollias20246th,kollias20247th}. In this work, we have focused on the task of estimating valence and arousal for the 8th ABAW competition \cite{kolliasadvancements}. In particular, for the valence-arousal track, the dataset is provided with 594 videos of around 2,993,081 frames obtained from 584 subjects. Sixteen of these videos display two subjects, both of which have been annotated. The annotations for valence and arousal are provided continuously in the range of $\lbrack-1,1\rbrack$. The dataset is divided into training, validation, and test sets of 356, 76, and 162 videos respectively. The partitioning is done in a subject-independent manner so that every subject’s data will be present in only one subset. 

\subsection{Evaluation Metric}
The Concordance Correlation Coefficient (CCC) is the widely used evaluation metric in the literature of dimensional ER to measure the level of agreement between the predictions ($\widetilde y$) and ground truth ($y$) of valence and arousal. Let $\mu_{\widetilde y}$ and $\sigma_{\widetilde y}^2$ represent the mean and variance of predictions respectively. Similarly, $\mu_y$ and $\sigma_y^2$ denote the mean and variance of ground truth, respectively, the CCC between the predictions and ground truth can be obtained as
\begin{equation}
CCC=\frac{2\sigma_{{\widetilde y}y}^2}{\sigma_{\widetilde y}^2+\sigma_y^2+(\mu_{\widetilde y}-\mu_y)^2}
\end{equation}
where $\sigma_{{\widetilde y}y}^2$ denotes the covariance of predictions and ground truth. 

Although Mean Square Error (MSE) has been widely used as a loss function for regression models, CCC was found to be more relevant for continuous regression labels \cite{atmaja2021evaluation}. So we also used CCC-based loss function, following the literature of dimensional ER \cite{cite7, 10005783}, which is given by 
\begin{equation}
   L=\sum_{c\in\{v,a\}}(1-CCC(\widetilde y^c,y^c))
\end{equation}


\subsection{Implementation Details}
\subsubsection{Preprocessing}
For visual modality, we have used the cropped and aligned images provided by the challenge organizers \cite{kollias2021analysing}. All the faces are resized to $48$x$48$ and missing faces in the video frames are replaced as black frames (i.e., zero pixels). Some of the frames are not annotated and we discard those frames.
The video sequence is divided into sub-sequences of length $300$ (i.e., $L$=300) with a stride $200$ frames, resulting in 33\% overlap, thereby providing 33\% more data. 

For audio modality, the audio stream is extracted from the videos and sampled with a sampling rate of 16KHz. Now log melspectrograms are computed from the sampled audio stream using the code provided by vggish repository\footnote{https://github.com/harritaylor/torchvggish}. Note that the audio modality is properly synchronized with the corresponding subsequences of visual modality using a hop length of $1/fps$ of the raw videos, while extracting the spectrograms. 

\subsubsection{Training Details}
For visual modality, we have used Resnet-50 \cite{7780459} pretrained on MS-CELEB-1M dataset \cite{10.1007/978-3-319-46487-9_6}, which is further finetuned on FER+ dataset \cite{barsoum2016training}. 
For audio modality, we used VGG-Net architecture pretrained on large-scale audioset dataset \cite{45611}. For both audio and visual modalities, TCNs are used to capture the temporal dynamics of frame-level embeddings. Data augmentation is used for the visual modality, where random flipping and random crop of size 40 is used for training images, while only center crop is used for validation images. Both audio and visual features are normalized in order to have a mean and standard deviation of 0.5. The models are trained separately for valence and arousal. To regularize the network, a weight decay of $0.001$ is used with Adam optimizer. The batch size is set to be 12. To avoid overfitting, we have employed early stopping and the maximum number of epochs are set to be 100. The hyperparameters of the initial learning rate and minimum learning rate are set to be $1e-5$ and $1e-8$ respectively. In our training mechanism, we also deployed a warmup scheme using $ReduceLRonPlateau$ scheduler with a patience of 5 and a factor of 0.1 based on CCC score of the validation set. 
Inspired by the performance of \cite{10208757}, we also employ gradual finetuning of backbones, where three groups of layers for visual (Resnet-50) and audio (VGG) backbones are progressively selected for finetuning. More specifically, the first group is unfrozen at epoch 0 and the learning rate is linearly warmed up to $1e-5$ within an epoch. Then repetitive warm-up is employed until epoch 5, after which $ReduceLRonPlateau$ scheduler is used to update the learning rate. The learning rate is gradually dropped with a factor 0.1 until validation CCC does not improve over 5 consecutive epochs. Now the second group is fine-tuned and the learning rate is set to $1e-5$, followed by warm-up scheme with $ReduceLRonPlateau$. The process is repeated until all the layers are finetuned for audio and visual backbones. To mitigate the problem of overfitting, the best model state dictionary over prior epochs is loaded at the end of each epoch. We also employed 6-fold cross-validation, where fold-0 is the official partition of training and validation sets provided by the organizers \cite{kolliasadvancements}.    

\section{Results and Discussion}
\subsection{Ablation Study}
\begin{table*}
\setlength{\tabcolsep}{4pt}
\renewcommand{\arraystretch}{1.25}
    \centering
    \caption{ CCC of the proposed approach compared to state-of-the-art methods for multimodal fusion on the original Affwild2 validation set (fold 0). ${\star}$ indicates that results are presented with the implementation of our experimental setup. Highest scores are shown in bold.}
    \label{tab:comparisiontoSOA}
    \begin{tabular}{|c|c|c|c|c|c|c|c|c|c|c|} 
	\hline
	\textbf{Method}& 
 \textbf{Type of } & \multicolumn{2}{|c|}{\textbf{Validation Set}} & \multicolumn{2}{|c|}{\textbf{Test Set}} \\
    \cline{3-6}
     & \textbf{Fusion} & \textbf{Valence} & \textbf{Arousal} & \textbf{Valence} & \textbf{Arousal} \\
\hline \hline
        Zhang et al. \cite{9607460} & Leader-Follower & 0.469 & 0.649 & 0.463 & 0.492\\
  \hline
    Zhang et al. \cite{10208713} & Transformers & 0.464 & 0.640 & \textbf{0.648} & 0.625\\
  \hline

  Zhou et al. \cite{10209026} & Transformers & 0.550 & \textbf{0.681} & 0.500 & \textbf{0.632}\\
  \hline
  Zhang et al. \cite{10208807} & Transformers & 0.554 & 0.659 & 0.523 & 0.545\\
  \hline
  Meng et al. \cite{Situ-RUCAIM3} & Transformers & 0.588 & 0.668 & 0.606 & 0.596\\
  \hline
  
 Praveen et al \cite{Praveen_2022_CVPR} & JCA  & 0.663 & 0.584 & 0.374 & 0.363\\ 
	\hline
 	Praveen et al \cite{10095234}$^{\star}$ & RJCA  & 0.443 & 0.639 & 0.537  & 0.576\\ 
	\hline
     Praveen et al \cite{10687371}$^{\star}$ & DCA  & 0.451 & 0.647 & 0.549  & 0.585\\ 
	\hline
	GRJCA (Ours) & GRJCA  & 0.459 & 0.652 & 0.556 & 0.605 \\ 
	\hline
    	HGRJCA (Ours) & HGRJCA  & 0.464 & 0.660 & 0.561 & 0.620 \\ 
	\hline
\end{tabular}
\end{table*}
\begin{table*}
\label{ablation}
\renewcommand{\arraystretch}{1.25}
    \centering
    \caption{ CCC of the proposed fusion model by varying the number of recursions. Bold indicates the highest scores. Fold 1 is used for experiments with multiple recursions.}
    \label{Ablation Study}
    \begin{tabular}{|c|c|c|c|c|c|c||c|c|c|c|} 
	\hline
	 \textbf{Number of } & \multicolumn{3}{|c|}{\textbf{Valence}}  &  \multicolumn{3}{|c|}{\textbf{Arousal}} \\
     \cline{2-7}
     \textbf{recursions ($t$)} & \textbf{RJCA} &\textbf{GRJCA} & \textbf{HGRJCA} & \textbf{RJCA} & \textbf{GRJCA} & \textbf{HGRJCA} \\
	\hline
	$l$ = 1  & 0.571 & 0.585 & 0.591 &  0.649 & 0.658 & 0.660 \\
	\hline
    $l$ = 2 & 0.575 & 0.593 & 0.606 & 0.653 & 0.664 & 0.669 \\
	\hline
	$l$ = 3 & 0.582 & 0.601 & \textbf{0.623} & 0.659 & 0.668 & \textbf{0.671}\\
	\hline
	$l$ = 4 & 0.580 & 0.596 & 0.617 & 0.652 & 0.661  & 0.662\\
	\hline
\end{tabular}
\end{table*}

Table \ref{Ablation Study} presents the results of the experiments conducted on the validation set to analyze the impact of the number of recursions of the RJCA model with the proposed GRJCA and HGRJCA models. First, we conducted a series of experiments with our baseline fusion model RJCA model by varying the number of recursions. Initially, we implemented the RJCA model with a single recursion, which is same as Joint Cross-attention model \cite{10005783}. Now the number of recursions are gradually increased and found that the performance of the system also improves gradually, achieving best results at 3 iterations. 
Now, we conducted another series of experiments to understand the impact of the GRJCA model by varying the number of recursions. Though RJCA model helps in obtaining more refined feature representations, they fail to deal with weak complementary relationships. By introducing GRJCA model, we can observe that the relative performance has been consistently improved over that of RJCA model as we increase the number of iterations. This demonstrates that handling weak complementary relationships across audio and visual modalities plays a key role in effectively modeling the synergic inter-modal relationships. Similar to that of RJCA model, we achieved best results for GRJCA model with 3 iterations. Beyond that, the system performance declines, which can be attributed to over-fitting. Finally, we conducted another series of experiments with HGRJCA model by varying the number of recursions. We can observe that HGRJCA model offers slight improvement in performance by applying the gating mechanism in a hierarchical fashion. We hypothesize that employing gating mechanism at the individual iterations followed by high level gating helps to capture more relevant information with finer granularity.


\subsection{Comparison to state-of-the-art}
Table \ref{tab:comparisiontoSOA} shows the performance of the proposed approach against the relevant state-of-the-art audio-visual fusion models on the official validation set of the Affwild2 dataset. Most of the related work on the Affwild2 dataset has been submitted to the Affective Analysis in-the-wild (ABAW) challenges \cite{kollias2021analysing,kollias2022abaw}. Therefore we compare our proposed approach with the relevant state-of-the-art models of ABAW challenges for A-V fusion in dimensional ER. One the strategies employed by majority of these approaches to improve the generalization ability is to increase the training dataset by leveraging additional datasets. Some of these approaches also explored multiple backbones for each modality to obtain diverse feature representations to improve the performance on test set \cite{9857097,10209026}. Though leveraging additional datasets and exploring multiple architectures with ensembling improves the performance, it is often computationally expensive and cumbersome. 
Zhang et al \cite{9607460} 
proposed a leader-follower attention mechanism by considering the visual modality as the primary channel, while the audio modality is used as a supplementary channel to improve the fusion performance. 
Zhang et al. \cite{10208713} showed that employing Masked-autoencoders on the visual modality, followed by fusion with audio modality can achieve better generalization and showed consistent improvement on both valence and arousal. 
Praveen et al. \cite{10005783} introduced joint cross attention model to encode the intra-modal relationships along with the inter-modal relationships simultaneously, achieving significant improvement in the validation set, especially for valence. They further improved their approach by introducing recursive mechanism into the joint cross attention framework, and demonstrated that recursive fusion helps in obtaining more refined feature representations. However, all these methods overlook the problem of weak complementary relationships, assuming audio and visual modalities always complement with each other. Praveen et al. \cite{10687371} investigated this problem and proposed DCA model to handle the weak complementary relationships by controlling the flow of information between original unattended and final attended features. 
In this work, we can observe that the proposed GRJCA model further improves the performance over DCA model by employing gating mechanism on the attended features of all the iterations. We also showed further improvement with HGRJCA model by employing gating mechanism in a hierarchical fashion, which helps to control the flow of subtle information across the iterations at fine-grained level. Note that even though the proposed models show lower performance on the official validation set, they achieved better performance eon other folds of cross-validation set, which is used to generate test set predictions. 



\begin{table}
\renewcommand{\arraystretch}{1.25}
    \centering
    \caption{ CCC of the proposed approach on Affwild2 test set compared to other methods submitted to 8th ABAW competition. Bold indicates best results of our approach on the test set.} 
    \label{Comparison with state-of-the-art for Affwild2 test}
    \begin{tabular}{|l|c|c|c|c|c|c||c|c|c|} %
	\hline
	 \textbf{Method }
	 & \textbf{Valence}  &  \textbf{Arousal} & \textbf{Mean}\\
	 \hline  \hline
	USTC-IAT-United \cite{yu2025interactive}   &   0.577 & 0.623 & 0.600 \\
	\hline
    GRJCA (Ours)   & \textbf{0.561} & \textbf{0.620} &  \textbf{0.590}\\
	\hline
	CtyunAI \cite{zhou2025emotion}  & 0.546 & 0.611 & 0.578 \\
	\hline
   HSEmotion \cite{savchenko2025hsemotion}  & 0.494 & 0.551 &  0.522 \\
	\hline
    AIWELL-UOC \cite{cabacas2024enhancing}  & 0.468 & 0.492 &  0.480 \\
	\hline
    Charon \cite{liang2025mamba}   & 0.504 & 0.412 &  0.458 \\
	\hline
        CAS-MAIS   & 0.327 & 0.304 &  0.316 \\
        \hline
    Baseline \cite{kollias20246th}   & 0.211 & 0.191 &  0.201 \\
	\hline
\end{tabular}
\end{table}

\subsection{Results on Test Set}
We have implemented 6-fold cross validation on both GRJCA and HGRJCA models and chose the best performing models on the validation set for generating the test set predictions. All the approaches that surpassed the baseline performance on the valence-arousal challenge of the 8th ABAW competition \cite{kolliasadvancements} are based on visual modality, while three approaches leveraged both audio and visual modalities. USTC-IAT-United \cite{yu2025interactive} also explored Resnet-50 and VGGish backbone for visual and audio modalities similar to our approach, whereas they have improved the temporal modeling of the modalities using multiscale TCNs, followed by cross-modal transformers. It is worth mentioning that the performance of the winner is close to our performance as they also employ similar backbones with different fusion models. CtyunAI \cite{zhou2025emotion} and AIWELL-UOC \cite{cabacas2024enhancing} explored CLIP encoder to improve the visual feature representations and showed that CLIP encoders can help to achieve more generalized visual feature representations. Unlike other approaches, the proposed approach tackles the under-explored problem of weak complementary relationships across audio and visual modalities. We have shown that handling weak complementary relationships using a simple gating mechanism to control the flow of information across the attended features of multiple iterations and unattended features can significantly improve the performance of the system.

\section{Conclusion}
In this paper, we have introduced two audio-visual fusion models with RJCA as a baseline fusion model using a simple, yet efficient gating mechanism to effectively capture the inter-modal relationships by handling weak complementary relationships of audio and visual modalities. Even though several approaches have been proposed for dimensional Emotion Recognition (ER) based on cross-modal interactions, the problem of weak complementary relationships is less explored in the literature. Although cross-attention based approaches have shown significant improvement in the fusion performance, we have shown that deploying a gated attention model further improves the performance of the system. Specifically, the proposed approach emphasizes the most relevant features by leveraging the most semantic information across the original features as well as the attended features of all iterations of the RJCA model. Extensive experiments conducted on the challenging Affwild2 dataset demonstrate the robust performanc eof the proposed approach. We have shown that the problem of weak complementary relationships is a promising line of reaseach, that can foster the performance of the model by effectively modliung the synergic relationships across the audio and visual modalities for emotion recognition. 


{
    \small
    \bibliographystyle{ieeenat_fullname}
    \bibliography{main}
}


\end{document}